\documentclass{article} % For LaTeX2e
\usepackage[final]{colm2026_conference}
\usepackage{amsmath}
\usepackage{graphicx}

\usepackage{microtype}
\usepackage{hyperref}
\usepackage{url}
\usepackage{booktabs}

\usepackage{lineno}

\definecolor{darkblue}{rgb}{0, 0, 0.5}
\hypersetup{colorlinks=true, citecolor=darkblue, linkcolor=darkblue, urlcolor=darkblue}

\title{Voice of Reason: Reinforcement Learning for Spoken Math}

\author{%
Timothée Weisselberger \\
Kyutai \\
Paris, France \\
\texttt{timothee.weisselberger@kyutai.org}
\And
Edouard Grave \\
Kyutai \\
Paris, France
\And
Alexandre Défossez \\
Kyutai, Gradium \\
Paris, France \\
\texttt{alex@kyutai.org}
}
\begin{document}

\ifcolmsubmission
\linenumbers
\fi

\maketitle

\begin{abstract}
Speech language models enable richer spoken interactions between humans and machines than cascaded systems, allowing access to paralinguistic information and lower latency. However, their accuracy on mathematical reasoning benchmarks has lagged behind those of text models.
Reinforcement learning (RL) with verifiable rewards has been instrumental in extending text models' capabilities for solving complex problems and limiting hallucinations.
In this work, we explore applying RL to the GLM-4-Voice speech model~\citep{glm4voice} to bridge the gap between textual and spoken mathematical problem solving. We first adapt the model to the domain using supervised fine-tuning on synthesized spoken question-answering data. We then show that, even without extra reasoning tokens, RL improves the accuracy on GSM8K beyond levels previously achieved for speech models only with supplementary reasoning traces. When combined with existing streaming reasoning techniques, we show further gains to 74.8\% free-form accuracy. This establishes a new state-of-the-art for mathematical spoken abilities with speech-native models.

% Reasoning in speech language models remains underexplored, especially in interleaved audio-text architectures where rewards are most naturally defined from text while generation depends on both text and audio-token decisions. In this work, we study reinforcement learning for spoken reasoning in GLM-4-Voice. We consider two training settings: a standard spoken-answer setting without explicit reasoning-token supervision in the output stream, and a STITCH-style setting in which the model is first fine-tuned to produce chunked reasoning traces before further reinforcement learning. In both cases, rewards are computed from the decoded text stream using an external language-model judge. We train on spoken reasoning examples derived from a filtered and regenerated version of the Orca math dataset with English TTS supervision. Our results show that reinforcement learning substantially improves spoken reasoning performance in GLM-4-Voice in both training settings. In particular, our best models achieve state-of-the-art results for spoken mathematical reasoning with GLM-4-Voice, highlighting the effectiveness of combining supervised fine-tuning with reinforcement learning in spoken reasoning pipelines.
\end{abstract}
\section{Introduction}

Text-based language models have seen a steep improvement in their problem-solving abilities, thanks to reasoning and reinforcement learning (RL) with verifiable rewards~\citep{shao2024deepseekmath}.
Various approaches have been tried to bring such advances to speech language models. 
Despite recent progress, the most effictive approach for building speech conversational models, when considering only intelligence and reasoning abilities, remains the cascading of speech-to-text, text-only, and text-to-speech models~\citep{chen2024voicebench}. 

The cascaded approach however suffers from the cumulative latency of each sub-model, and cannot access paralinguistic information. To alleviate these limitations, speech models with built-in audio support were developed. Some use light adapter layers around existing text models~\citep{wang2024freeze}, optionally with fine-tuning of the whole model~\citep{xu2025qwen25omni,glm4voice}. Conversational support is added using explicit turn boundary detection, i.e., using a voice activity detection model. On the other hand, full-duplex models were developed, either without any text stream~\citep{nguyen2023generative}, or with a text stream reflecting the current speech production of the model~\citep{moshi,roy2026personaplex}. Full-duplex models offer unrivaled latency; however, their performance on complex tasks falls short of cascaded or adapted text models~\citep{chen2024voicebench}.

There are several challenges when applying outcome-based reinforcement learning to speech models. First, existing datasets require some reformatting effort to be adapted to speech. Mathematical phrasing can be ambiguous or unnatural when spoken out. Such data can only be economically generated at scale through text-to-speech models, adding noise to the training data when such models fail. Second, speech models are constrained by real-time requirements and are expected to remain interactive at all times. Thus, audio tokens must be generated at regular intervals, and supplementary reasoning tokens must be kept within a reasonable limit~\citep{stitch}, especially for on-device applications.

% Improving reasoning ability in audio language models is still an underexplored area of Machine Learning despite these model being able to sustain spoken interaction. Reinforcement learning (RL) has become a standard in post-training text-only models for reasoning capabilities. Applying such methods to audio models directly involves some challenges. The reward being more naturally defined only on text tokens.

% This mismatch between text and audio tokens is especially acute in interleaved architectures such as GLM-4-Voice, where generation alternates between text and audio blocks. A naive policy-gradient objective assigns credit to many token-level audio decisions whose exact identity may have little direct effect on answer correctness, thereby introducing unnecessary variance. At the same time, audio tokens cannot simply be removed from the RL loss, since some of them still affect the structure and termination of the response.

In this work, we study RL for mathematical problem solving with speech language models, with a primary focus on the GLM-4-Voice model~\citep{glm4voice}. It exhibits strong base performance on math benchmarks, and was shown to improve with explicit reasoning-token supervision interleaved with the output stream~\citep{stitch}. We contribute by showing that even without reasoning tokens, the model can improve to unprecedented accuracy on GSM8K~\citep{gsm8k}, through in-domain fine-tuning followed by reinforcement learning with AI feedback. Moreover, our approach can be combined with the previous reasoning-token supervision method and yield state-of-the-art precision on this benchmark.

% Our goal is to improve reasoning performance without requiring explicit reasoning-token supervision in the output stream. We also study a complementary setting in which the model is first fine-tuned with STITCH-style chunked reasoning supervision, and then further improved with reinforcement learning. This lets us test whether RL remains beneficial even when explicit reasoning traces are already present in the output format.

% We address this problem with merged audio tokens, a grouped objective for audio positions. Instead of optimizing the log-probability of the exact sampled audio token, we optimize the total probability mass assigned to the set of audio tokens. Under a value-invariance assumption, this yields an unbiased lower-variance estimator.

% Empirically, we find that RL substantially improves reasoning performance in GLM-4-Voice without explicit reasoning tokens. We also observe that the  resulting pipeline is simpler than expected: supervised fine-tuning before RL is important, while several additional components, such as KL regularization or auxiliary audio-maintenance losses, appear less critical in this setting. Overall, our results suggest that effective RL for audio reasoning does not require a heavy pipeline, but rather a small number of targeted modifications that account for the structure of interleaved audio-text generation.

\section{Related Work}
\label{sec:related_work}

\textbf{Speech language models.}
Speech language models aim to produce speech with low latency while preserving semantic correctness (instruction-following, reasoning) and speech quality (naturalness, speaker consistency). They rely on discrete tokens provided by neural audio codecs~\citep{soundstream,encodec} which are modeled autoregressively~\citep{borsos2023audiolm}.
While initial research focused on unsupervised audio modeling, these models were adapted to interactive use through three approaches: (i) adapting existing text models with a light audio encoder mapping to the input space of a text language model, with a streaming TTS plugged to its output text stream~\citep{xu2025qwen25omni,wang2024freeze}; (ii) interleaving short blocks of either only text or only speech tokens, directly modeled by the backbone~\citep{nguyen2025spirit,glm4voice}; (iii) full-duplex models jointly modeling two audio streams, one for the user and one for the model, along with optional parallel text streams~\citep{nguyen2023generative,moshi,roy2026personaplex}. For (i) and (ii), turn changes between the user and assistant are indicated through the use of special delimiter tokens, whose insertion is triggered by voice activity detection models for end of turn prediction~\citep{wang2024freeze}. Despite architectural differences, all paradigms enforce tight text-speech alignment to keep generation streamable.

\textbf{Abilities of speech models.}
There are several axes along which speech language models can be evaluated.
When looking at the naturalness of the interaction and the turn-taking dynamics, they have been steadily improving~\citep{lin2025full,roy2026personaplex}. However, when measuring capabilities requiring deeper thinking and understanding, speech language models exhibit worse zero-shot abilities than equivalent text models, and require more training data in order to reach acceptable performance~\citep{moshi,zhang2025mimo}. In a number of speech-based tasks, cascaded systems wrapping existing text models with speech-to-text and text-to-speech models outperform speech-native models~\citep{chen2024voicebench}. 
Meanwhile, text models have solved increasingly complex tasks, such as the GSM8K benchmark~\citep{gsm8k}, with an accuracy of up to 91.6\% for a 7-billion-parameter model~\citep{qwen2025qwen25technicalreport}. For comparison, GLM-4-Voice~\citep{glm4voice} achieves 27.3\% accuracy with 9 billion parameters, which \citet{stitch} improves to 58.7\% with the addition of blocks of supplementary reasoning tokens in their STITCH method. As we show in Section~\ref{sec:results}, full-duplex models lag further behind on this benchmark.

\textbf{Alignment of speech models.}
While text models already exhibit strong problem-solving ability after pre-training~\citep{brown2020language}, they can be further improved through
reinforcement-learning-based alignment using either human preferences~\citep{ziegler2019finetuning}, or verifiable rewards~\citep{shao2024deepseekmath}, especially when combined with reasoning tokens.
When applied to the speech domain, reasoning tokens must be carefully interleaved with speech tokens, so as not to delay the audio output and preserve interactivity~\citep{stitch}.
RL for speech has previously been studied to improve factuality~\citep{wu2025aligning} and naturalness~\citep{speechjudge}, as well as for unsupervised simultaneous speech translation~\citep{labiausse2026simultaneous}. To the best of our knowledge, the present work is the first application of RL for mathematical reasoning in speech-native models.

\section{Method}

\begin{figure}[t]
    \centering
    \includegraphics[width=\linewidth]{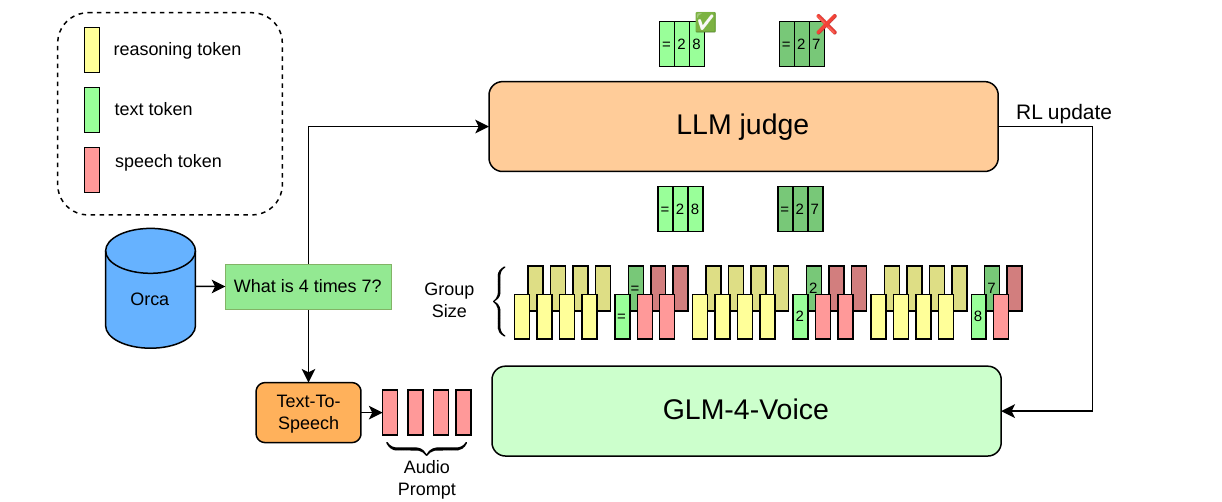}
    \caption{Illustration of our RL pipeline. We synthesize as speech a question taken from the Orca dataset~\citep{orcamath} to be fed to the GLM-4-Voice~\citep{glm4voice}, and generate multiple interleaved speech-text replies. We apply a group-relative policy-gradient objective inspired by GRPO~\citep{shao2024deepseekmath}, using a binary reward obtained from a large language model applied to the decoded text stream. The model can optionally use interleaved reasoning tokens following~\citet{stitch}.}
    \label{fig:our_figure}
\end{figure}

\subsection{Problem setup}
\label{sec:setup}

We study post-training for GLM-4-Voice \citep{glm4voice}, an interleaved spoken language model that generates text and audio in a fixed alternation pattern. It predicts auto-regressively one token at a time, belonging to either the text vocabulary $\mathcal{V}_{\text{text}}$, the audio vocabulary $\mathcal{V}_{\text{aud}}$, or a small set of special control tokens $\mathcal{V}_{\text{spec}}$.
% We denote the model by $\pi_\theta$. 
Given an audio-only input prompt of length $S$, $x \in \mathcal{V}_{\text{aud}}^S$, the model produces an assistant response of length $T$,
\[
y = (y_1,\dots,y_T),
\]
where each token belongs to either the text vocabulary $\mathcal{V}_{\text{text}}$, the audio vocabulary $\mathcal{V}_{\text{aud}}$, or a small set of special control tokens $\mathcal{V}_{\text{spec}}$.

Assistant responses in the training data follow a structured output format. This structure is learned rather than enforced by a hard vocabulary mask: under standard decoding, text and audio are expected to be emitted in alternating blocks following the native GLM-4-Voice interleaving format. In our setup, each standard spoken-response block consists of $13$ text tokens followed by $26$ audio tokens. Thus, if we ignore special markers, a typical response has the form
\begin{equation}
\label{eq:structure_base}
(\underbrace{t_1,\dots,t_{13}}_{\text{text}},
 \underbrace{a_1,\dots,a_{26}}_{\text{audio}},
 \underbrace{t_{14},\dots,t_{26}}_{\text{text}},
 \underbrace{a_{27},\dots,a_{52}}_{\text{audio}},
 \dots).
\end{equation}
Special tokens are additionally used for control purposes, for instance to delimit turns between the user and the model. Given that it takes more audio tokens than text tokens to represent a reply, once all text tokens are generated, the model will produce only audio tokens until a special end of turn token.

We also experiment with STITCH~\citep{stitch}, which inserts reasoning tokens learnt through direct supervision, to extend the thinking capabilities of the model.
%In the STITCH-style setting \citep{stitch}, illustrated in in Figure~\ref{fig:stitch_figure}, the model is additionally supervised to produce an explicit reasoning trace before each spoken-response block. 
Reasoning is emitted in successive chunks of $100$ text tokens, delimited by special markers, while the spoken response itself still follows the native GLM-4-Voice interleaving pattern. 
\citet{stitch} introduced two variants: STITCH-R and STITCH-S, depending on whether the first reasoning packet is before (-R) or after (-S) the first text-audio block. In the following, we focus on STITCH-R, and denote it STITCH.
A typical sequence therefore takes the form
\begin{equation}
\label{eq:structure_stitch}
(\underbrace{r_1,\dots,r_{100}}_{\text{reasoning text}},
 \underbrace{t_1,\dots,t_{13}}_{\text{text}},
 \underbrace{a_1,\dots,a_{26}}_{\text{audio}},
 \underbrace{r_{101},\dots,r_{200}}_{\text{reasoning text}},
 \underbrace{t_{14},\dots,t_{26}}_{\text{text}},
 \underbrace{a_{27},\dots,a_{52}}_{\text{audio}},
 \dots).
\end{equation}
Special tokens such as \texttt{[SOPR]}, \texttt{[EOPR]}, and \texttt{[EOR]} are inserted to delimit reasoning chunks and mark the end of the reasoning phase.

Our goal is to improve problem-solving accuracy of the GLM-4-Voice model with RL, whether reasoning tokens are used or not. 
% To evaluate a sampled response, we decode its text stream and score it with an external language-model judge. 

%The reward therefore depends primarily on the textual content of the answer, even though the policy generates both text and audio tokens and must respect the interleaved output format.

% This creates a structural mismatch. The reward is defined at the sequence level and mostly reflects textual correctness, while the policy-gradient estimator is computed over token-level actions, including many individual audio-token choices. These choices are numerous because every $13$ text tokens are followed by $26$ audio tokens, so a substantial fraction of the generated actions lie in the audio vocabulary even though the reward is mainly determined by the decoded text. The main question is therefore how to adapt policy-gradient training to this structured interleaved setting when the supervision signal is primarily text based.

\subsection{Group-relative policy gradient with AI feedback}
\label{sec:GRPO}

We assume a set of audio prompts corresponding to a given question or problem.
For each prompt $x$, we sample $G$ completions by sampling autoregressively from the model.
We denote by $\pi_\theta(y\mid x)$ the distribution over the output sequences given the audio prompt. Namely, we sample
\[
y^{(1)}, \dots, y^{(G)} \sim \pi_\theta(\cdot \mid x),
\]
and score each completion independently with a scalar reward
\[
r^{(g)} = R(x, y^{(g)}).
\]

We then form a group-relative baseline by centering rewards within the prompt:
\[
b = \frac{1}{G} \sum_{g=1}^G r^{(g)}, 
\qquad
A^{(g)} = r^{(g)} - b.
\]

Our basic RL objective is an on-policy policy-gradient loss of the form
\begin{equation}
\label{eq:grpo_base}
\mathcal{L}_{\mathrm{RL}}(\theta)
=
-\frac{1}{G}\sum_{g=1}^G
A^{(g)}
\sum_{t \in \mathcal{I}^{(g)}}
\log \pi_\theta\!\left(y_t^{(g)} \mid y_{<t}^{(g)}, x\right),
\end{equation}
where $\mathcal{I}^{(g)}$ denotes the set of generated positions included in the loss. In practice, this corresponds to a group-relative REINFORCE objective~\citep{williams1992simple}. It is related to GRPO~\citep{shao2024deepseekmath} but we do not use PPO clipping.

The reward is derived from an LLM judge where the judge returns a binary correctness score for the decoded text response. We sample multiple completions per prompt with stochastic decoding and score them independently with the judge model. The reward is computed from the decoded text stream only. A representation of the overall modeling and training pipeline is provided in Figure~\ref{fig:our_figure}.

\subsection{Temperature correction}
\label{sec:temp_correction}

During RL, trajectories are sampled with stochastic decoding using temperature $T < 1$. Therefore, the policy that generates the samples is not the base model distribution $\pi_\theta$, but the temperature-adjusted policy
\[
\pi_\theta^{(T)}(y_t \mid s_t)
=
\frac{\exp(z_\theta(y_t, s_t)/T)}
{\sum_{y'\in\mathcal{V}} \exp(z_\theta(y',s_t)/T)},
\]
where $z_\theta(\cdot,s_t)$ denotes the model logits at decoding state $s_t=(x,y_{<t})$.

To keep the policy-gradient estimator consistent with the actual sampling distribution, we compute log-probabilities under $\pi_\theta^{(T)}$ rather than under the raw model distribution. In practice, this simply means dividing logits by $T$ before applying the log-softmax in the RL loss, following prior RLHF practice \citep{ziegler2019finetuning}.

Accordingly, the objective \eqref{eq:grpo_base} becomes
\begin{equation}
\label{eq:grpo_temp}
\mathcal{L}_{\mathrm{RL}}(\theta)
=
-\frac{1}{G}\sum_{g=1}^G
A^{(g)}
\sum_{t \in \mathcal{I}^{(g)}}
\log \pi_\theta^{(T)}\!\left(y_t^{(g)} \mid y_{<t}^{(g)}, x\right).
\end{equation}

Without this temperature correction, trajectories are sampled from one distribution but optimized under another, which introduces a mismatch in the gradient estimator. As shown by the ablations in Section~\ref{sec:results}, this correction is critical in our setting.

% \subsection{Why text-based reward is problematic in interleaved generation}

% If the reward depended equally on all generated tokens, the loss above would be a natural choice. In our setting, however, the reward is computed from the decoded text stream. As a result, many audio-token choices contribute variance to the estimator while having limited direct influence on reward.

% A natural alternative would be to exclude audio tokens from the RL loss and optimize only over text tokens. For interleaved models such as GLM-4-Voice, this is too crude. Audio-side decisions still affect the temporal structure of generation and, in particular, the way the response terminates. Audio tokens therefore cannot simply be discarded from the objective.

% This motivates a coarser treatment of audio actions: we would like to keep the relevant decision that the model follows an audio branch, while removing distinctions between individual audio tokens whenever those distinctions do not affect reward.

\subsection{Structure of the vocabulary and loss calculation}
\label{sec:token_merging}

The structure of the output space of the GLM-4-Voice model raises interesting questions. 
% First, practical use of the model requires for the audio tokens to remain consistent with the text tokens.
Given that the reward model only verifies the correctness of the output text, nothing
prevents the GLM-4-Voice model from departing from the structured output pattern described in eq. \eqref{eq:structure_base} and eq. \eqref{eq:structure_stitch}. When applying the RL objective given by eq. \eqref{eq:grpo_temp}, we can decide whether to apply it only on text tokens, or both on audio and text tokens. In particular, as the audio tokens are strongly constrained by the preceding text tokens, it could be sufficient to only apply the RL loss to the text outputs. Yet, as described in Section~\ref{sec:setup}, the end of the text reply is implicitly indicated by the presence of an audio token in place of a text one. 

This motivated us to test a third alternative, where we collapse the probabilities over all audio tokens into a single abstract audio token $a^*$. Formally, we introduce a new policy
\[
\bar{\pi}_\theta^{(T)}(y_t = a^* \mid s_t)
=
\sum_{a\in\mathcal{V}_{\text{aud}}}\pi_\theta^{(T)}(y_t = a \mid s_t),
\]
to be used in place of $\pi_\theta^{(T)}$ in \eqref{eq:grpo_temp}.
This yields another estimator which is under some assumptions lower-variance and unbiased; the formal statement and proof are given in Appendix \ref{app:maths}. Unless stated otherwise, this setup is used in the experiments.

% We noticed that the only contribution of audio tokens to the reward (discarding their auto-regressive impact) is by the fact that there is an audio token or not. So we derived an objective where we merge all audio-tokens into one to reduce variance in the training.
% Let $\mathcal{V}_{\text{aud}}$ denote the set of audio tokens. For a generated sequence $y$, define $s_t = (x, y_{<t})$ as the decoding state at position $t$. Under the standard objective, an audio position contributes
% \[
% \log \pi_\theta(y_t \mid s_t),
% \qquad y_t \in \mathcal{V}_{\text{aud}}.
% \]

% With audio tokens merging, we replace this term by the log-probability mass assigned to the whole audio-token set:
% \[
% \log \pi_\theta(\mathcal{V}_{\text{aud}} \mid s_t)
% =
% \log \sum_{v \in \mathcal{V}_{\text{aud}}} \pi_\theta(v \mid s_t).
% \]

% Equivalently, the token-level contribution becomes
% \[
% \ell_t(\theta) =
% \begin{cases}
% \log \pi_\theta(y_t \mid s_t), & \text{if } y_t \in \mathcal{V}_{\text{text}}, \\[6pt]
% \log \pi_\theta(\mathcal{V}_{\text{aud}} \mid s_t), & \text{if } y_t \in \mathcal{V}_{\text{aud}}.
% \end{cases}
% \]

% The resulting loss is
% \[
% \mathcal{L}_{\mathrm{merge}}(\theta)
% =
% -\frac{1}{G}\sum_{g=1}^G
% A^{(g)}
% \sum_{t \in \mathcal{I}^{(g)}}
% \ell_t^{(g)}(\theta).
% \]

% The intuitive idea is to remove small distinctions between audio tokens that do not matter (the reasoning is happening in the text tokens). This yields another estimator which is under some assumptions lower-variance and unbiased; the formal statement and proof are given in Appendix \ref{app:maths}.

\section{Experimental setup}

\subsection{Models}
\label{sec:models}
Our main experiments are conducted on GLM-4-Voice~\citep{glm4voice}, an interleaved audio-text language model that alternates text-token blocks and audio-token blocks during generation. All reported results start from the publicly available checkpoint \footnote{\href{https://huggingface.co/zai-org/glm-4-voice-9b}{huggingface.co/zai-org/glm-4-voice-9b}}. We do not modify the model architecture and study only post-training effects. We release our flagship reasoning-free and STITCH-style checkpoints as
\href{https://huggingface.co/kyutai/glm-4-voice-of-reason-9b}
{\texttt{glm-4-voice-of-reason-9b}} and
\href{https://huggingface.co/kyutai/glm-4-voice-of-reason-stitch-9b}
{\texttt{glm-4-voice-of-reason-stitch-9b}}, respectively.

\subsection{Training data}
\label{sec:training_data}
Our training data is derived from a filtered and regenerated version of the Orca-Math dataset \citep{orcamath} which has been shown not to be contaminated by GSM8K. It consists of question-answer pairs on problems requiring basic mathematical thinking. 
We reformulate both questions and answers with a text large language
model, Qwen 3 235B~\citep{yang2025qwen3technicalreport}, so that they
can be synthesized as speech by the DSM text-to-speech
model~\citep{zeghidour2025streaming}. Training audio is generated
using many different DSM voices. For evaluation, we reuse the audio
provided by the STITCH authors, which was generated with
GPT-4o-mini-TTS. Thus, the training and evaluation audio are generated
with different TTS systems. We derive three datasets to be used for post-training, either for
in-domain supervised fine-tuning (SFT) or RL.

\textbf{Reasoning-free SFT.} We derive interleaved text and audio tokens following the format given by eq.~\eqref{eq:structure_base} in Section~\ref{sec:setup} to be used as direct supervision for the model output when input with the speech tokens of the question.

\textbf{Reasoning STITCH-like SFT.} We follow the STITCH methodology of ~\citet{stitch}. A reasoning trace is first generated to arrive at the answer, which is then chunked, each chunk being summarized in a speech-compatible manner using again a text model. The concatenation of all summaries is synthesized to speech. This gives us interleaved reasoning, text, and speech tokens as layout in eq.~\ref{eq:structure_stitch}, in Section~\ref{sec:setup}, to be used as direct supervision.

\textbf{Reinforcement learning.} For reinforcement learning, we only need the question speech tokens, to be fed as prompt, and the question text, fed to the text judge described in Section~\ref{sec:GRPO}.

Each dataset is derived from the same 150,616 samples from the Orca corpus after filtering out those with a combined speech and text token length exceeding 2,000 (without reasoning).

\subsection{Reward model}
\label{sec:reward_model}

During RL, each sampled completion is scored by an external language-model judge applied to the decoded text stream. The judge receives the input question text together with the decoded assistant response and returns a binary reward for correctness, as shown in Figure~\ref{fig:our_figure}. The prompt passed to the judge is provided in Appendix~\ref{app:judge_prompt}.

Our training judge is
\texttt{qwen/qwen3-235b-a22b-2507}~\citep{yang2025qwen3technicalreport}.
The training judge does not receive the reference answer.

We manually inspect 100 training judgments. The judge agrees with
human annotations in 88\% of cases, with 80.0\% precision and 95.2\%
recall for correct answers. Its errors are predominantly false
positives, indicating that the reward is somewhat permissive but
remains strongly correlated with correctness.

We separately inspect 100 judgments produced by the evaluation judge,
which receives the reference answer, and observe agreement with human
annotations on all 100 examples.

\subsection{Training procedure}
\label{sec:training_procedure}

We post-train both a reasoning-free GLM-4-Voice~\citep{glm4voice} model, as well as the STITCH-style variant.

\textbf{SFT.}  For both, we start with supervised fine-tuning on the datasets described in Section~\ref{sec:training_data} using a standard cross-entropy loss on the reply tokens.
We perform the SFT on the entire dataset, except in Table~\ref{tab:ft_quantity} where we experiment with using a subset of the training data. We start from the publicly available GLM-4-Voice checkpoint. 
Note that there is no publicly available STITCH checkpoint. Thus, we fine-tune the first checkpoint, first without reasoning traces, and then with reasoning traces on the entire Orca-derived dataset.

\textbf{RL.}
For the default RL stage, we sample 4 completions per question using
stochastic decoding with a temperature of 0.9 and a maximum generation
length of 600 tokens for the reasoning-free setting and 800 tokens for
the STITCH-style setting. We optimize the objective given by
Eq.~\ref{eq:grpo_temp} in Section~\ref{sec:temp_correction}.

In Table~\ref{tab:ft_quantity}, we additionally report longer
reasoning-free RL runs using a group size of 8. These longer runs are
individual runs rather than multi-seed estimates.

\textbf{Optimization.}
We use AdamW~\citep{adamw} with a batch size of 16, a weight decay of $0.1$, $\beta_1=0.9$ and $\beta_2=0.95$. For SFT, we use a learning rate of $2\cdot10^{-6}$ and train for one
epoch; for RL, we use a learning rate of $10^{-7}$ and perform
1,500 updates. All experiments run on 16 H100 GPUs.

% Importantly, our RL pipeline does not use KL regularization. We found it to have little impact while causing instability in training.

% Our default setup is thus: standard spoken-answer supervision, binary reward, merging of audio tokens, group size 4, temperature 0.9 and no KL regularization.

% Additional implementation details, including exact decoding settings, judge prompts, and dataset preprocessing, are deferred to the appendix. \ref{}

\subsection{Evaluation benchmarks and metrics}
\label{sec:evaluation}

\paragraph{Mathematical reasoning.}
We use GSM8K~\citep{gsm8k} as our main mathematical reasoning benchmark. The model receives each question as speech and produces an interleaved text--audio response.

Following \citet{stitch}, for text-stream evaluation we provide the original question, the decoded text stream generated by the model, and the ground-truth answer to a text-model judge tasked with returning a binary correctness score. We use GPT-4o through the OpenAI API%
\footnote{\href{https://developers.openai.com/api/docs/models/gpt-4o}{developers.openai.com/api/docs/models/gpt-4o}},
along with the prompt from the Kimi Audio EvalKit~\citep{kimiteam2025kimiaudio}%
\footnote{\href{https://github.com/MoonshotAI/Kimi-Audio-Evalkit/blob/a65bd4243dd280daa685e35a5cfd69c371315c06/almeval/datasets/ds_refqa.py\#L9}{github.com/MoonshotAI/Kimi-Audio-Evalkit/almeval/datasets/ds\_refqa.py}}.

\paragraph{Spoken-output evaluation.}
To verify that improvements in the decoded text stream are reflected
in the generated speech, we transcribe the generated audio with \texttt{Qwen/Qwen3-ASR-1.7B}~\citep{qwen3asr} and apply the same final-answer
evaluation to the ASR transcript. We evaluate speech naturalness with
UTMOSv2~\citep{baba2024utmosv2} on 100 generated GSM8K responses. ASR is used only for evaluation and is not part of the reward in our main experiments.

\paragraph{Out-of-domain evaluation.}
To measure whether mathematical post-training degrades more general capabilities, we evaluate the models on a spoken 1,000-example subset of
TriviaQA~\citep{joshi2017triviaqa}. This benchmark is not used during mathematical SFT or RL.

\paragraph{Response characteristics.}
We report the average number of generated spoken-text tokens, audio tokens, hidden reasoning tokens, and the average duration of the spoken response. This allows us to determine whether improvements are explained by substantially longer generations or reasoning traces.

\paragraph{Statistical reporting.}
For our main results, we report the mean and standard deviation over three independent training seeds. The exact evaluation protocol, including dataset splits, decoding settings, checkpoint selection, and aggregation details, is provided in Appendix~\ref{app:external_eval}. We also investigated MultiArith~\citep{multiarith}, SingleEQ~\citep{singleeq}, SVAMP~\citep{svamp}, and AddSub~\citep{addsub}. However, a paraphrase-level analysis revealed substantial overlap between these benchmarks and the Orca-derived training corpus. We therefore remove them from the main evaluation and do not use them as evidence of generalization. The contamination analysis is reported in Appendix~\ref{app:contamination}.

\subsection{Baselines and ablations}

Our main point of comparison is the original STITCH model~\citep{stitch}. We provide both their reported numbers in Table~\ref{tab:external_baselines}, as well as the results of our re-implementation detailed in Section~\ref{sec:training_procedure}, in Table~\ref{tab:main_comparison}, allowing for an evaluation of our RL contribution free of biases from the change in training data. 

We also compare to the state-of-the-art full-duplex model PersonaPlex~\citep{roy2026personaplex} using its publicly available checkpoint, as well as to the Qwen2.5-Omni model~\citep{xu2025qwen25omni}. We additionally include Qwen3-Omni-30B~\citep{qwen3omni} and a cascaded ASR--LLM--TTS--ASR baseline using Gemma-4-31B-IT~\citep{gemmateam2026gemma4} as the text reasoning model and Kokoro~\citep{kokoro82m} as the speech synthesizer. The larger omni and cascaded systems are included as top lines rather than parameter-matched comparisons. % TODO: add the Qwen3-Omni citation.
We also provide ablation studies on the amount of SFT data used in Table~\ref{tab:ft_quantity}, and on the RL loss strategy in Table~\ref{tab:other_ablations}.
% Because training data and pipeline details differ across setups, we do not rely only on reported numbers; instead, we reproduce a STITCH-style setting on our own Orca-derived data by reusing their prompting scheme to construct chunked reasoning targets from the same question distribution.

% We study two training regimes:
% (i) a standard spoken-answer setting without explicit reasoning chunks,
% and (ii) a STITCH-style setting in which the model is first fine-tuned with chunked reasoning supervision and then further optimized with RL.

% Within these regimes, our main comparisons are between the base model, SFT only, and SFT followed by RL.

% We further study ablations of the RL objective, including audio token merging, KL regularization, reward design, group size, and same-data versus disjoint-data SFT/RL splits.

% ============================================================
% External baselines
% ============================================================

\begin{table}[t]
\centering
\small
\begin{tabular}{lllr}
\toprule
\textbf{Method}
& \textbf{\# Param.}
& \textbf{Output}
& \textbf{GSM8K} \\
\midrule

\multicolumn{4}{c}{\textit{Speech-native models}} \\
\midrule

PersonaPlex~{\citep{roy2026personaplex}}
    & 8B
    & Speech
    & 3.2 \\

GLM-4-Voice~{\citep{glm4voice}}
    & 9B
    & Speech
    & 27.3 \\

STITCH~{\citep{stitch}}
    & 9B
    & Speech
    & 58.7 \\

% GLM RL seeds:
% 1ea9a8c4/ckpt_100
% a756d739/ckpt_100
% b2035c15/ckpt_100
GLM-4-Voice + SFT + RL (Ours)
    & 9B
    & Speech
    & $65.5 \pm 1.1$ \\

% STITCH RL seeds:
% ce284eee/ckpt_100
% 94a60b69/ckpt_100
% 43cb5b86/ckpt_70
STITCH-style SFT + RL (Ours)
    & 9B
    & Speech
    & $\mathbf{74.8 \pm 1.1}$ \\

\midrule
\multicolumn{4}{c}{\textit{Omni and cascaded top lines}} \\
\midrule

Qwen2.5-Omni~{\citep{xu2025qwen25omni}}
    & 7B
    & Text
    & 84.7 \\

Qwen3-Omni~{\citep{qwen3omni}}
    & 30B
    & Text
    & 94.6 \\

Cascaded ASR--LLM--TTS--ASR
    & 31B LLM
    & Speech
    & $\mathbf{95.7}$ \\

\bottomrule
\end{tabular}

\caption{
Accuracy on GSM8K.
Results for our models are reported as mean and standard deviation
over three independent training seeds.
Qwen3-Omni and the cascaded system are stronger top lines, but are
not parameter- or architecture-matched to our speech-native models.
The cascaded system uses Gemma-4-31B-IT as the reasoning model and
Kokoro as the speech synthesizer.
}
\label{tab:external_baselines}
\end{table}

% ============================================================
% Main SFT/RL comparison:
% performance, speech evaluation, OOD evaluation and lengths
% ============================================================

\begin{table*}[t]
\centering
\small

\textbf{(a) Mathematical accuracy, spoken-output evaluation,
and OOD generalization}

\vspace{0.3em}

\resizebox{\textwidth}{!}{%
\begin{tabular}{lrrrr}
\toprule
\textbf{Method}
& \textbf{GSM8K text}
& \textbf{GSM8K ASR}
& \textbf{TriviaQA}
& \textbf{UTMOSv2} \\
\midrule

GLM-4-Voice base
    & 27.3
    & 26.4
    & $\mathbf{40.6}$
    & 3.614 \\

GLM-4-Voice + SFT
    & $61.7 \pm 1.9$
    & 61.5
    & $33.4 \pm 0.5$
    & 4.067 \\

GLM-4-Voice + SFT + RL
    & $\mathbf{65.5 \pm 1.1}$
    & $\mathbf{63.9}$
    & $34.0 \pm 2.0$
    & $\mathbf{4.069}$ \\

\midrule

STITCH-style SFT
    & $68.0 \pm 1.6$
    & $66.2 \pm 2.3$
    & $20.2 \pm 1.0$
    & $\mathbf{4.174}$ \\

STITCH-style SFT + RL
    & $\mathbf{74.8 \pm 1.1}$
    & $\mathbf{72.0 \pm 1.9}$
    & $\mathbf{21.4 \pm 2.2}$
    & 4.164 \\

\bottomrule
\end{tabular}
}

\vspace{0.8em}

\textbf{(b) Average response characteristics}

\vspace{0.3em}

\begin{tabular}{lrrrr}
\toprule
\textbf{Method}
& \textbf{Text tokens}
& \textbf{Audio tokens}
& \textbf{Duration}
& \textbf{Reasoning tokens} \\
\midrule

GLM-4-Voice base
    & 84
    & 419
    & 33.5\,s
    & 0 \\

GLM-4-Voice + SFT
    & 152
    & 524
    & 41.9\,s
    & 0 \\

GLM-4-Voice + SFT + RL
    & 130
    & 455
    & 36.4\,s
    & 0 \\

\midrule

STITCH-style SFT
    & 60
    & 184
    & 14.7\,s
    & 167 \\

STITCH-style SFT + RL
    & 80
    & 251
    & 20.0\,s
    & 176 \\

\bottomrule
\end{tabular}

\caption{
Impact of supervised domain adaptation and reinforcement learning.
Text-stream GSM8K accuracy is computed from the model's decoded text
tokens, while ASR accuracy is computed after transcribing the generated
speech and applying the same final-answer metric.
ASR is used only for evaluation.
UTMOSv2 is evaluated on 100 GSM8K responses.
TriviaQA results measure out-of-domain generalization on a spoken
1,000-example subset.
The STITCH-style SFT + RL TriviaQA result uses two seeds; the main
GSM8K results use three independent seeds.
}
\label{tab:main_comparison}
\end{table*}

% ============================================================
% Amount of SFT data
% ============================================================

\begin{table}[t]
\centering
\small
\begin{tabular}{lrr}
\toprule
\textbf{Method}
& \textbf{GSM8K}
& \textbf{TriviaQA} \\
\midrule

% SFT sig: f4293188
10\% SFT
    & 43.9
    & 39.4 \\

10\% SFT + RL
    & 50.5
    & $\mathbf{42.3}$ \\

% Long group-size-8 RL run:
% sig: 5361e106
10\% SFT + RL$^\dagger$
    & $\mathbf{58.5}$
    & 41.4 \\
\midrule

% SFT sig: 0862ee2b
50\% SFT
    & 56.4
    & $\mathbf{36.5}$ \\

% RL sig: 59fbd189
50\% SFT + RL
    & $\mathbf{61.5}$
    & 36.3 \\

\midrule

% SFT seeds:
% 629ba1f7/ckpt_60
% 0e804a5f/ckpt_60
% d4dd735d/ckpt_60
Full-data SFT
    & $61.7 \pm 1.9$
    & $33.4 \pm 0.5$ \\

% Default RL, three seeds:
% 1ea9a8c4/ckpt_100
% a756d739/ckpt_100
% b2035c15/ckpt_100
Full-data SFT + RL
    & $65.5 \pm 1.1$
    & $34.0 \pm 2.0$ \\

% Longer group-size-8 flagship:
% sig: 22840b41/ckpt_70
Full-data SFT + RL$^\dagger$
    & $\mathbf{70.8}$
    & $\mathbf{35.6}$ \\

\bottomrule
\end{tabular}

\raggedright
{
$^\dagger$ Longer individual RL run using a group size of 8;
no multi-seed uncertainty is reported.
}

\caption{
Impact of the amount of supervised fine-tuning data before RL.
Rows marked with $^\dagger$ correspond to longer individual RL runs using a group size of 8; all other RL runs use the default group size of 4. Non-dagger full-data results are reported as mean and standard deviation over independent training seeds. Dagger and low data rows are derived from individual training runs.
}
\label{tab:ft_quantity}
\end{table}

% ============================================================
% RL objective ablations
% ============================================================

% ============================================================
% RL objective ablations
% ============================================================

% ============================================================
% RL objective ablations
% ============================================================

\begin{table}[t]
\centering
\small
\begin{tabular}{llccr}
\toprule
\textbf{Method}
& \textbf{Loss support}
& \textbf{Temp. correction}
& \textbf{Group size}
& \textbf{GSM8K} \\
\midrule

Default
    & Audio-token merging
    & Yes
    & 4
    & $65.5 \pm 1.1$ \\

No temperature correction
    & Audio-token merging
    & No
    & 4
    & 12.3 \\

Loss on all tokens
    & All generated tokens
    & Yes
    & 4
    & 64.4 \\

Loss on text tokens
    & Text tokens only
    & Yes
    & 4
    & 63.8 \\

Larger group
    & Audio-token merging
    & Yes
    & 8
    & \textbf{67.0} \\

\bottomrule
\end{tabular}

\caption{
Ablation of the RL objective on GSM8K.
The default configuration uses audio-token merging, temperature
correction, and a group size of 4, and is reported as mean and standard
deviation over three independent training seeds.
The remaining results are averages over late-training checkpoints from
individual training runs.
}
\label{tab:other_ablations}
\end{table}

\section{Results}
\label{sec:results}

\textbf{Comparison to external baselines.}
We report in Table~\ref{tab:external_baselines} a comparison on GSM8K of our final RL-improved models against a number of state-of-the-art baselines. We first notice that, despite their improved interactive capabilities, full-duplex models such as PersonaPlex~\citep{roy2026personaplex} lag far behind turn-based ones.

GLM-4-Voice~\citep{glm4voice} acts as a stronger baseline, reaching 27.3\% on GSM8K~\citep{gsm8k}, while STITCH~\citep{stitch} reaches 58.7\% using reasoning steps, an improvement of 31.4 points. Our RL approach brings the reasoning-free GLM-4-Voice model from 27.3\% to $65.5 \pm 1.1$\%, a gain of 38.2 points, and surpasses the original STITCH model by 6.8 points without using explicit reasoning tokens. We further show that STITCH-style supervision and our RL approach can be combined, reaching $74.8 \pm 1.1$\%, the strongest result among the speech-native models considered here.\footnote{These results use the original decoding setup, which applies top-$k=50$. Removing this constraint increases GSM8K accuracy to 70.3\% for the reasoning-free model and 77.1\% for the STITCH-style model. The released checkpoints linked in Section~\ref{sec:models} correspond to these models.}

Qwen2.5-Omni, Qwen3-Omni, and the cascaded baseline reach 84.7\%, 94.6\%, and 95.7\%, respectively. These larger omni and cascaded systems remain strong top lines, although they are not strictly matched to our models in terms of size, architecture, and output constraints.

\textbf{Impact of our RL pipeline.}
We report in Table~\ref{tab:main_comparison} the impact of the SFT and RL stages described in Section~\ref{sec:training_procedure}. First, we notice the positive impact of in-domain SFT using a high-quality dataset such as Orca~\citep{orcamath}, which improves GLM-4-Voice from 27.3\% to $61.7 \pm 1.9$\% on GSM8K (+34.4 points). We further confirm the relevance of the STITCH approach of~\citet{stitch}. With our training data, STITCH-style supervision improves performance from $61.7 \pm 1.9$\% to $68.0 \pm 1.6$\% (+6.3 points). RL improves the reasoning-free model from $61.7 \pm 1.9$\% to $65.5 \pm 1.1$\% (+3.8 points), and improves the STITCH-style model from $68.0 \pm 1.6$\% to $74.8 \pm 1.1$\% (+6.8 points). Overall, RL
consistently closes part of the remaining gap after supervised training, while STITCH-style supervision provides a stronger starting point than standard spoken-answer supervision.

\textbf{Spoken-output evaluation.}
The improvements measured on the decoded text stream are also reflected in the generated speech. The base GLM-4-Voice model obtains 26.4\% ASR-based accuracy, close to its 27.3\% text-stream accuracy. For STITCH-style models, ASR-based GSM8K accuracy improves from $66.2 \pm 2.3$\% after SFT to $72.0 \pm 1.9$\% after RL. For reasoning-free GLM-4-Voice, ASR-based accuracy similarly improves from 61.5\% to 63.9\% (+2.4 points).

Speech naturalness improves substantially during SFT and remains stable
during RL: for reasoning-free GLM-4-Voice, UTMOSv2 increases from 3.614 for the base model to 4.067 after SFT and 4.069 after RL, while for STITCH-style models it changes from 4.174 to 4.164 after RL. Thus, the improvement is visible in the spoken output and does not come with a measurable degradation in naturalness.

\textbf{Response characteristics.}
Table~\ref{tab:main_comparison} also reports average response lengths and durations. For reasoning-free GLM-4-Voice, RL reduces the average response from 152 to 130 text tokens ($-22$), from 524 to 455 audio tokens ($-69$), and from 41.9 to 36.4 seconds ($-5.5$ seconds) relative to SFT. The improvement is therefore not explained by longer responses.

For STITCH-style models, RL increases the spoken response from 60 to 80 text tokens ($+20$) and from 14.7 to 20.0 seconds ($+5.3$ seconds). However, the number of reasoning tokens remains nearly unchanged, increasing only from 167 to 176 ($+9$). The accuracy gain is therefore not obtained through
substantially longer reasoning traces.

\textbf{Out-of-domain generalization.}
We evaluate the models on a spoken 1,000-example subset of TriviaQA. The original GLM-4-Voice model reaches 40.6\%. Full-data SFT reduces this score to $33.4 \pm 0.5$\% ($-7.2$ points), while adding RL increases it to $34.0 \pm 2.0$\% (+0.6 points over SFT).

Similarly, adding RL to the STITCH-style model improves TriviaQA from $20.2 \pm 1.0$\% to $21.4 \pm 2.2$\% (+1.2 points). In contrast, using only 10\% of the SFT data largely preserves general capabilities: TriviaQA accuracy increases from 39.4\% after SFT to 42.3\% after RL (+2.9 points). In the 50\% regime, it changes only from 36.5\% after SFT to 36.3\% after RL ($-0.2$ points). The longer group-size-8 runs obtain 41.4\% in the 10\% regime and 35.6\% in the full-data regime. These results suggest that the observed out-of-domain degradation mainly comes from specialization during full-data SFT rather than from RL.

\textbf{Amount of SFT training.}
Table~\ref{tab:ft_quantity} shows that  Performance increases steadily with more supervised data: 43.9\% with 10\% SFT, 56.4\% with 50\% SFT, and $61.7 \pm 1.9$\% on the full dataset. This indicates that reasoning
quality strongly depends on supervised initialization.

RL consistently improves GSM8K on top of every SFT regime. With the
default group size of 4, performance improves from 43.9\% to 50.5\%
in the 10\% regime (+6.6 points), from 56.4\% to 61.5\% in the 50\%
regime (+5.1 points), and from $61.7 \pm 1.9$\% to $65.5 \pm 1.1$\% in the full-data regime (+3.8 points). Relative gains are larger when SFT data is limited, suggesting that RL is particularly valuable in weaker-data regimes, while the strongest reasoning-free performance is obtained from
full-data SFT followed by longer RL training. Most strikingly, with only 10\% of the supervised data, the longer RL
run reaches 58.5\%, gaining 14.6 points over SFT alone and recovering over 80\% of the gap to full-data SFT. It comes within just 3.2 points of the full-data SFT model, showing that RL can replace much of the benefit of a tenfold increase in supervised data.

\textbf{Design of the RL loss.}
We next study ablations on the RL objective, starting from our default setup: binary reward, audio-token merging
(see Section~\ref{sec:token_merging}), group size 4, temperature 0.9, and no KL regularization. We vary four components: whether temperature correction is applied in the policy-gradient estimator
(Section~\ref{sec:temp_correction}; \textit{no temperature correction}), whether the loss is computed on all generated tokens instead of using audio-token merging (\textit{all tokens}), whether optimization is restricted to text tokens only (\textit{text only}), and whether the group size is increased from 4 to 8 (\textit{group size 8}).
Table~\ref{tab:other_ablations} reports the resulting performance. The default result is reported as mean and standard deviation over three independent training seeds. The remaining results are averages over late-training checkpoints from individual runs.

The most striking result is the importance of temperature correction.
Removing it causes a dramatic collapse, dropping GSM8K from
$65.5 \pm 1.1$\% to 12.3\%. The other variants remain closer to
the default configuration: the \textit{all tokens} variant reaches 64.4\%, the
\textit{text only} variant reaches 63.8\%, and increasing group size from
4 to 8 gives the best ablation result at 67.0\%.
\section{Conclusion}

We show that reinforcement learning significantly improves spoken
mathematical reasoning in speech-native models. Starting from GLM-4-Voice, a simple post-training pipeline yields large gains in the standard spoken-answer setting, without requiring explicit reasoning tokens, reaching $65.5 \pm 1.1$\% on GSM8K. When combined with STITCH-style supervision, RL further improves accuracy to $74.8 \pm 1.1$\%, the strongest result among the speech-native models considered in this work. The trade-off between using additional reasoning tokens and reducing inference cost depends
on the application; in particular, the simpler reasoning-free model may be valuable for power-efficient on-device inference. Overall, our results show that reinforcement learning is a powerful
tool for closing part of the gap between spoken and text-based reasoning while preserving streaming-compatible generation and speech naturalness. Further work is required to extend these methods to full-duplex models and to combine best-in-class interactivity and
naturalness with reasoning capabilities approaching those of top-line
text and cascaded systems.

\bibliography{arxiv_clean}
\bibliographystyle{colm2026_conference}

\clearpage
\appendix

\section{Prompt templates}
\label{app:prompts}

In this appendix, we report the main prompt templates used in our experiments. For readability, we show them in lightly edited form, with variable placeholders written in braces.

\subsection{Training reward judge prompt}
\label{app:judge_prompt}

\begin{quote}
\small
\begin{verbatim}
You are an evaluator for question/answer pairs.
The user's message is a reasoning question.

Return Answer: 1 ONLY if the assistant response is correct and answers the question.
Otherwise return Answer: 0.

Write your output in EXACTLY two lines:
1) "Reasoning: <at most {budget} words>"
2) "Answer: X" where X is EXACTLY one of: 0, 1
Do NOT add anything else.

User question:
"""
{user_input}
"""

Assistant response:
"""
{assistant_response}
"""
\end{verbatim}
\end{quote}

\section{RL loss estimator under audio token-merging}
\label{app:maths}

\paragraph{Exact Rao--Blackwellization under value invariance.}

Let \(y_t\) denote the token emitted at step \(t\), and let \(a^*\) denote the
abstract audio event, i.e. the event that the emitted token belongs to the
audio vocabulary:
\[
y_t = a^*
\quad\Longleftrightarrow\quad
y_t \in \mathcal V_{\mathrm{aud}}.
\]
We define the corresponding abstract probability mass as
\[
\bar{\pi}_\theta^{(T)}(y_t = a^* \mid s_t)
:=
\sum_{a\in\mathcal V_{\mathrm{aud}}}
\pi_\theta^{(T)}(y_t=a\mid s_t).
\]

Consider the standard policy-gradient form
\[
\nabla_\theta J(\theta)
=
\mathbb E\!\left[
\sum_t A_t\,\nabla_\theta \log \pi_\theta^{(T)}(y_t\mid s_t)
\right],
\]
where \(A_t\) is any valid advantage estimator.
For a fixed state \(s_t\), define
\[
Q(s_t,a)
=
\mathbb{E}\left[
R(x,y)
\mid s_t,\ y_t=a
\right].
\]
Assume that on audio steps, the value is invariant to the identity of the
audio token:
\begin{equation}
Q(s_t,a)=Q_{\mathrm{aud}}(s_t)
\qquad
\text{for all } a\in\mathcal V_{\mathrm{aud}}.
\label{eq:audio_invariance_assumption}
\end{equation}
Then the within-event Rao--Blackwell identity holds:
\begin{align}
\mathbb E\!\left[
Q(s_t,y_t)\,\nabla_\theta \log \pi_\theta^{(T)}(y_t\mid s_t)
\,\middle|\, s_t,\, y_t=a^*
\right]
&=
Q_{\mathrm{aud}}(s_t)\,
\mathbb E\!\left[
\nabla_\theta \log \pi_\theta^{(T)}(y_t\mid s_t)
\,\middle|\, s_t,\, y_t=a^*
\right]
\nonumber\\
&=
Q_{\mathrm{aud}}(s_t)\,
\nabla_\theta \log \bar{\pi}_\theta^{(T)}(y_t=a^*\mid s_t).
\label{eq:rb_exact}
\end{align}
Indeed,
\begin{align*}
\mathbb E\!\left[
\nabla_\theta \log \pi_\theta^{(T)}(y_t\mid s_t)
\,\middle|\, s_t,\, y_t=a^*
\right]
&=
\sum_{a\in\mathcal V_{\mathrm{aud}}}
\frac{\pi_\theta^{(T)}(y_t=a\mid s_t)}
{\bar{\pi}_\theta^{(T)}(y_t=a^*\mid s_t)}
\,\nabla_\theta \log \pi_\theta^{(T)}(y_t=a\mid s_t)\\
&=
\nabla_\theta \log \bar{\pi}_\theta^{(T)}(y_t=a^*\mid s_t).
\end{align*}

Thus, under \eqref{eq:audio_invariance_assumption}, replacing the token-level score \(\nabla_\theta \log \pi_\theta^{(T)}(y_t\mid s_t)\) by the abstract-event score \(\nabla_\theta \log \bar{\pi}_\theta^{(T)}(y_t=a^*\mid s_t)\) on audio positions yields an unbiased gradient estimator. Moreover, since this replacement is a conditional expectation (Rao--Blackwellization), it cannot increase variance:
\[
\mathrm{Var}\!\left(\mathbb E[g\mid s_t,\, y_t=a^*]\right)\le \mathrm{Var}(g),
\qquad
g = Q(s_t,y_t)\nabla_\theta \log \pi_\theta^{(T)}(y_t\mid s_t).
\]

\section{Evaluation details}
\label{app:external_eval}

\paragraph{Evaluation audio.}
For GSM8K, we reuse the evaluation audio released by the STITCH authors, generated with GPT-4o-mini-TTS. This TTS system differs from DSM, which is used to generate our training data. Inputs are provided to all speech models as audio only.

For TriviaQA, we evaluate on a fixed spoken subset of 1,000 examples.

\paragraph{Text-stream accuracy.}
For GSM8K and TriviaQA, we evaluate the decoded text stream using GPT-4o as a binary final-answer judge. The judge receives the original question, the generated response, and the reference answer. We use the prompt from the Kimi Audio EvalKit, set the judge temperature to zero, use a reasoning budget of 150 words, and parse the final binary decision from its output.

For our models, inference is performed with temperature zero. We allow up to 800 generated tokens in the reasoning-free setting and 1,300 tokens in the STITCH-style setting.

\textbf{Spoken-output accuracy.}
We use Qwen3-ASR-1.7B~\citep{qwen3asr} to transcribe the generated speech. We then apply the same GPT-4o
final-answer evaluation to the ASR transcript. ASR is used only for evaluation and is never used as a training reward.

\paragraph{Speech naturalness.}
We evaluate speech naturalness with UTMOSv2~\citep{baba2024utmosv2} on 100 generated GSM8K responses for each reported model. The same evaluation set is used
before and after RL.

\paragraph{Statistical reporting.}
The main text-stream GSM8K results are reported as mean and standard deviation across three independent training seeds. The STITCH-style SFT + RL TriviaQA result uses two independent seeds. Results displayed without uncertainty are point estimates from the available evaluated runs.

For each training run, we use the final checkpoint of each run for multi-seed reporting; the additional late checkpoints are evaluated only for the individual-run experiments described below.

For the low-data and controlled-ablation experiments, only one
training run is available. When several late checkpoints are
evaluated, we report their mean without interpreting checkpoint
variation as variance across independent runs.

\paragraph{External baselines.}
For PersonaPlex, inference is performed with seed 42424242. To allow
the model to finish its response, we append silence to the input
waveform by repeating a silence clip 30 times. We read the generated
text from the model's JSON output and evaluate it with the same
final-answer judge.

Qwen2.5-Omni and Qwen3-Omni are evaluated using their text outputs.
Their responses are scored using the same GSM8K final-answer
evaluation as our models.

The cascaded baseline follows an
ASR--LLM--TTS--ASR pipeline. It uses Gemma-4-31B-IT as the text
reasoning model and Kokoro as the speech synthesizer. The final
transcript is evaluated using the same GSM8K final-answer judge.

\section{Contamination analysis}
\label{app:contamination}

We investigate paraphrase-level overlap between the Orca-Math
corpus~\citep{orcamath} used to construct our post-training data and four small arithmetic benchmarks initially considered for evaluation: AddSub~\citep{addsub}, MultiArith~\citep{multiarith}, SingleEQ~\citep{singleeq}, and SVAMP~\citep{svamp}. These benchmarks were included in the STITCH evaluation suite~\citep{stitch}, which motivated us to examine them in our setting.
For each evaluation question, we first retrieve candidate Orca-Math examples having the same final short answer. This inexpensive filtering step substantially reduces the number of candidate pairs. We then use \texttt{google/gemma-4-31b-it}~\citep{gemmateam2026gemma4} to determine whether each
candidate pair is a reformulation of the same underlying mathematical problem. An evaluation question is marked as contaminated when at least one Orca-Math example is judged to express the same underlying
problem.

Table~\ref{tab:contamination} reports the number of contaminated
questions among the evaluated questions for which the analysis was
performed.

\begin{table}[h]
\centering
\small
\begin{tabular}{lrrr}
\toprule
\textbf{Benchmark}
& \textbf{Questions}
& \textbf{Contaminated}
& \textbf{Ratio} \\
\midrule
AddSub
    & 108
    & 80
    & 74.1\% \\

MultiArith
    & 174
    & 82
    & 47.1\% \\

SingleEQ
    & 109
    & 102
    & 93.6\% \\

SVAMP
    & 287
    & 102
    & 35.5\% \\

\midrule
Total
    & 678
    & 366
    & 54.0\% \\
\bottomrule
\end{tabular}

\caption{
Paraphrase-level contamination between Orca-Math and the arithmetic
evaluation benchmarks. Candidate pairs are first filtered by matching
their final short answers and are then classified by
\texttt{google/gemma-4-31b-it}.
}
\label{tab:contamination}
\end{table}

We observe substantial overlap for all four benchmarks, ranging from
35.5\% on SVAMP to 93.6\% on SingleEQ. We therefore exclude these
benchmarks from the main results and do not use them as evidence of
generalization. Our mathematical reasoning evaluation instead focuses
on GSM8K, while out-of-domain preservation is studied separately on
the spoken TriviaQA subset.
% \section{Dataset Sizes and Splits}
% \label{app:dataset_splits}

% All training stages use the same Orca-derived filtered dataset. In particular, standard supervised fine-tuning (SFT), STITCH-style fine-tuning, and reinforcement learning (RL) are all performed on the same training set rather than on separate or disjoint subsets.

% The dataset contains exactly 150,616 samples. It is constructed from an Orca-derived corpus after filtering out examples whose combined question, generated responses and synthesized audio exceed 2000 tokens.

% Therefore, unless otherwise specified, the data used for SFT, STITCH-style fine-tuning, and RL are identical across experiments. No separate RL-only split is introduced in the settings considered in this work.

\end{document}